# Estimation of Soft Robotic Bladder Compression for Smart Helmets using IR Range Finding and Hall Effect Magnetic Sensing

Colin Pollard, Jon Aston, and Mark A. Minor[1], IEEE Member

*Abstract—* This research focuses on soft robotic bladders that are used to monitor and control the interaction between a user's head and the shell of a Smart Helmet. Compression of these bladders determines impact dissipation; hence the focus of this paper is sensing and estimation of bladder compression. An IR rangefinder-based solution is evaluated using regression techniques as well as a Neural Network to estimate bladder compression. A Hall-Effect (HE) magnetic sensing system is also examined where HE sensors embedded in the base of the bladder sense the position of a magnet in the top of the bladder. The paper presents the HE sensor array, signal processing of HE voltage data, and then a Neural Network (NN) for predicting bladder compression. Efficacy of different training data sets on NN performance is studied. Different NN configurations are examined to determine a configuration that provides accurate estimates with as few nodes as possible. Different bladder compression profiles are evaluated to characterize IR range finding and HE based techniques in application scenarios.

## I. Introduction

This paper examines techniques to estimate bladder compression in a soft robotic helmet, termed the "Smart Helmet", Fig. 1 (top) [1]. This helmet system consists of a hard shell covered with smart pneumatic bladders along its inner surface in place of traditional helmet padding. These bladders exert a wide range of forces on the head during impact. Being able to measure the deflection of the bladders is crucial for characterizing impacts and controlling the helmet. Space is very constrained, however, and as such, bladders in [1] utilized a valve, pressure sensor, and IR range finder embedded in the silicone bladder base, Fig. 1 (bottom). This paper augments that instrumentation with hall-effect sensors as an alternative for measuring bladder deflection. The goal of this paper is to examine how these different sensing techniques can characterize bladder deflection.

This research faces challenges due to the nature of the bladders and the helmet. The bladders are very flexible and have a large area in which they deflect, with an internal height of 30 mm and a radius of 17 mm. Compared to other soft robotic applications [2, 3] this compliance results in relatively large deflections. Further, bladder wall deflection during compression can obscure the line of sight between the sensor and the top of the bladder, potentially providing false readings with range finders.

Previous implementations of the helmet system [1] were designed to solely utilize an IR range finder to estimate the position, $Z_T$, of the top of the bladder, Fig. 1. A logarithmic model is applied to linearize the sensor data, but since the regression still has variations throughout its range, multiple techniques are examined to further correct discrepancies. The first segments the regression into multiple regions and applies complement filter to fuse the regions. A Neural Network (NN) is also examined for estimating bladder compression distance from range finder data. This paper presents the first published calibration and characterization of the range finder for sensing deflection of these bladders.

As a result, we also examined hall-effect sensors, which measure the presence of magnetic fields, and unlike optical sensors, do not require a clear line of sight. By embedding a small magnet in the top of the bladder and embedding an array of hall effect sensors in the bladder base, it becomes possible to estimate not just the compression, $Z_T$, but also lateral and

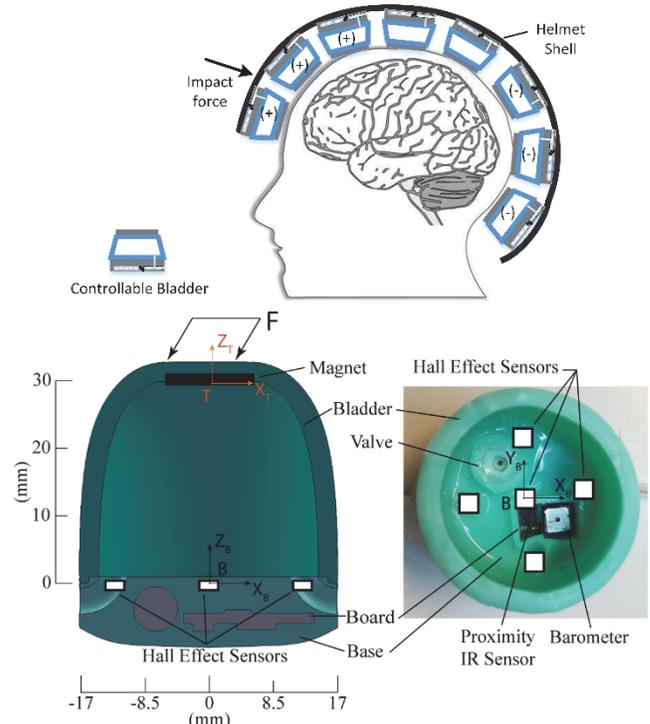

Fig. 1 Smart Helmet [1] (top) with embedded bladders. Detailed bladder design (bottom) with added range finder, hall effect sensors, and magnet.

*Research supported by the National Science Foundation under grant No. 1622741.

C. Pollard, J. Aston, and M. Minor are with the University of Utah, Salt Lake City, UT 84112 USA. (Email:colin.pollard1466@gmail.com, mark.minor@utah.edu)

longitudinal deflections $X_T$ and $Y_T$, respectively. This additional data is important to future iterations of Smart Helmet design concerning shear forces experienced by the bladders during off-centered impacts. The challenge, however, is that space is relatively limited and there is insufficient space for a large sensing array typical of other soft robotic solutions [4-6]. Hence, this paper contributes spatial sensing of relatively large displacements for soft robotics in compact spaces.

Another challenge is that location of the sensors within the bladder base is uncertain. Variations in sensor location on the PCB and location of the board during manufacturing naturally have small variations. The base of the bladder also naturally deflects during compression. Localizing magnets typically requires significant dimensional accuracy to solve complex mathematical models [7], which is not possible here. This study proposes a NN to transform Hall-Effect data into 3D positional data. Additionally, it demonstrates that it is possible to obtain accurate data when mounting sensors in a flexible substrate that actively deforms during bladder compression. Several parameters of the system are compared, including the neural network size, filtering constraints, training patterns, and angle sensitivity. Again, this paper contributes the first application of hall effect sensing in the Smart Helmet bladders.

The paper first discusses related work in Sec II, followed by a description of the bladder system in Sec III. IR range finding techniques are then presented in Sec IV and HE sensing techniques are in Sec V. Conclusions and future work are in Sec VI.

## II. Related Work

Soft robotic applications have utilized a number of different sensing techniques that are suited to compliant environments and actuation. Magnetic sensing lends itself well to this application due to its robustness and lack of dependence on line-of-sight. These properties have been utilized in a number of ways such as tactile force sensing in soft skins [3] and sensors [2, 8], as well as position/orientation localization for soft robots [4, 6]. These works however, either utilize large scale magnetic sensing arrays, or only are utilized to measure relatively small deflections. The intended application of this work intends to measure larger deflections on the scale of 20-30 mm in a compact space that only allows for a small array of sensors.

Optical based sensors such as cameras and photoelectric sensors are commonly seen in soft robotics applications. Cameras have been utilized to track pins embedded in a deformable surface to characterize the surface displacement during loading [9]. IR sensing was used in an application with similar form factor measuring the $Z$ deflection of a pressurized actuator array [10]. For this application these sensing techniques were avoided as camera imaging is affected by occlusions and is too computationally heavy for microcontroller implementation. IR sensing is examined in this paper as a way to measure the vertical compression of the bladder. Data-driven calibration techniques such as [11] that utilize regression-based fits were utilized to calibrate IR sensor data.

In order to estimate position from magnetic sensing data several processing techniques can be utilized. Model-based methods, such as [7], typically solve complex mathematical models. However, due to the material properties of our bladder substrate, sensors have noticeable differences in position and orientation during creation of evaluation prototypes. Bladder bases also undergo a non-trivial amount of compression and flexion during use. These two issues can cause conflicts with model-based methods due to uncertainty with sensor position.

Neural Networks are also applied for position estimation. In soft sensing applications, NN have been used with magnetic sensors extensively, showing the ability to obtain low-error localization approximations with as low as 4 magnetic sensors [3, 5, 6, 12]. Due to the complexity of characterizing magnetic fields, the neural network approach is well posed to our application. This research examines application of NN to fit both IR range finder and HE data to estimate deflection of the Smart Helmet bladders. In the case of the later, the paper contributes a study of how NN parameters effect displacement estimates.

## III. System Description

### A. Bladder Design

The bladders were designed to collapse through the first mode of buckling. Bladders have a wide base and taper slightly toward the top, where it is rounded off, Fig. 1. The bladders are created using 3D printed molds, into which silicon Mold Max 40 is then poured. The bladders are created in two parts: the bladder shell, and the bladder base that houses the mechatronics. The parts are then joined to create an enclosed system using the same silicon. The base was designed to include passive tubing connection between bladders but was sealed for the purposes of this study.

### B. Embedded Mechatronics

In the base of each bladder there is a 2 x 2 cm printed circuit board containing a microcontroller, pressure sensor, accelerometer, and a VCNL4000 IR rangefinder. Embedded in the rubber next to the circuit board is a small electric solenoid valve that can vent atmospheric air into and out of the bladder.

Embedded in the base above the circuit board and solenoid are five Allegro A1326 HE sensors. These analog ratiometric sensors output a voltage from 0-5 Volts. Each sensor is attached to an external non-inverting op-amp with a gain of 22V/V, an RC low-pass filter with a cutoff frequency of 100Hz, and a 10-bit analog to digital converter. The arrangement of Hall effect sensors is important to capture the motion of the magnet as it moves around the workspace. HE sensors are not capable of picking up magnetic fields when the fields are parallel to the face of the sensor, thus the spacing of sensors must be wide enough to avoid these conditions at high lateral deflections. To reduce complexity, we decided on a 5 single-axis HE sensor arrangement, Fig. 1. Other researchers have used more sensors to cover a wider area [13], but only 5 were used here due to the small are to place sensors.

Finally, a thin cylindrical magnet is embedded in the top of the bladder. The magnet size was chosen based on two factors. First, it had to physically fit inside of the rubber walls, and not impede compression of the bladder. Second, the strength of the magnet – primarily affected by the diameter and grade – was chosen such that at maximum compression, only the sensors directly under the magnet were saturated (output 5V), and the other sensors were within the operating voltages.

## IV. Rangefinder Evaluation

To sense bladder deflection, an IR rangefinder is embedded into the base of the bladder on a printed circuit board. It is oriented vertically and is mounted off-center with respect to the center of the bladder. The rangefinder outputs dimensionless counts which can be linearized using a logarithmic model. To calibrate the rangefinder, a centered, vertical compression is used where linearized counts are recorded with respect to known vertical displacement. These counts can then be converted to a distance. The methods demonstrated here are a piecewise regression, a complementary filtered regression, and a NN. The bladder is compressed through seven test trajectories designed to mimic helmet impacts and the predicted output is compared with the recorded output from the manipulation system. These trajectories are described further in the next Section, V.B.

### A. Piecewise Linear Regression

According to manufacturer specification, the outputted counts should follow a logarithmic scale and operate in a linear region while in the 2 to 100mm range. As such, a logarithmic model was fit to the sensor using a regression. In our testing, however, the logarithmic output functions in a linear region at close distances and has a well-defined point at which the model no longer correlates. This effect can be observed in Fig. 2, where at counts lower than 4200, the noise in the system goes up dramatically. For this reason, two linear regression models were fitted to the output of the logarithmic fit. The first operates in the region of 1 to 18mm, and the second from 18 to 25mm. To combine the two regression models, a piecewise function was implemented that outputs the respective regression depending on if the inputted counts is above or below the crossover. The function is described below where x is equal to counts:

$$\text{distance}(x) = \begin{cases} -0.03049 * x + 143.2, & x < 4100 \\ -0.01226 * x + 66.3, & x \geq 4100 \end{cases}$$

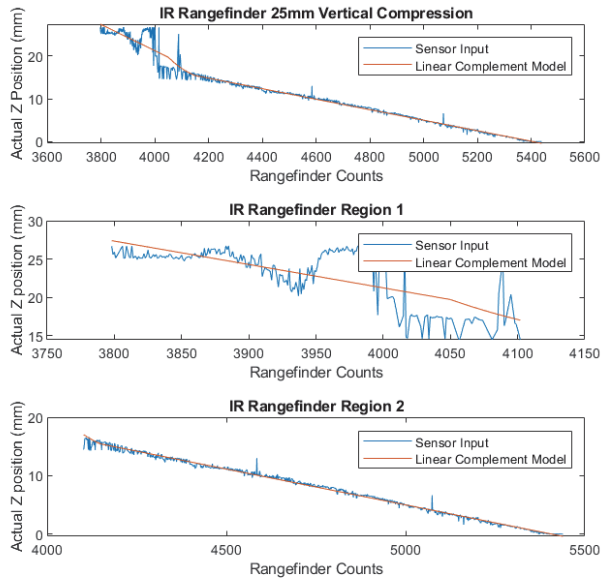

Fig. 2 Collected data from IR sensor (top) and linear fit (bottom)

### B. Complementary Linear Regression

An alternative method to fusing the two linear regression models is to utilize a complementary filter. In this configuration, each linear regression is weighted by the function α(counts), where counts is the reading from the rangefinder. The points at which the filter transitions from one region to the next were determined by examining the vertical compression trial and estimating the region in which the least confidence in each regression exists. In this case, the region existed in a space of 100 points between 4150 counts and 4050 counts. The function is described below:

$$\text{distance}(x) = \alpha(x) * z_1(x) + (1 - \alpha(x)) * z_2(x)$$

where,

$$z_1(x) = -0.03049 * x + 143.2$$
$$z_2(x) = -0.01226 * x + 66.3$$

and:

$$\alpha(x) = \begin{cases} 0, & x < 4050 \\ 1, & x > 4150 \\ \dfrac{x - 4050}{4150 - 4050}, & otherwise \end{cases}$$

### C. Neural Network

The rangefinder inside the bladder experiences several non-linear effects. These include imperfect reflection due to the silicon construction of the bladder, to non-perpendicular reflection surfaces from lateral bladder deflections. In the worst case, lateral deflection of the bladder can result in the rangefinder losing line of sight with the top of the bladder. For these reasons, it is desirable to improve the linear regression model to account for these effects. One method explored here is the use of a neural network.

A two-layer feed-forward network was trained on a repeated offset compression pattern. This pattern is described further later in this paper when comparing the effects of training data. The first layer contained 10 hidden neurons, while the second layer contained 1. The number of neurons was chosen based on empirical hand-tuning of the network. The training set consisted of approximately 110,000 samples. Training was accomplished using Bayesian-Regularization as it was found to produce better error compared to a Levenberg-Marquardt method.

## D. IR Range Finder Results

As shown in Table 1, the piecewise linear regression and complementary linear regression performed similarly. Both of these techniques encountered increased error in non-vertical tests. In these tests, there is a longer portion of the compression where line-of-sight can be lost through buckling due to bladder geometry. These effects are most pronounced on the offset Tests 4 through 7 and are maximized on Tests 4 and 7 where the bladder is deflected in the positive X and Y directions. This asymmetry in the positive versus negative offset directions could be due to the offset rangefinder position in the bladder base. In contrast, the neural network does not perform as well at vertical trials compared to the linear regression models; however, it does account for more effects in the non-vertical tests 2 through 7. Unfortunately, none of the three methods were able to overcome the noise introduced at distances over 18mm from the rangefinder, and as such the overall RMSE is much higher than the specification, further demonstrating the difficulty of sensing in the bladder environment at these distances.

Table 1. Range finder calibration results for different compression testing profiles.

|  | Piecewise RMSE ± STDEV (mm) | Complement RMSE ± STDEV (mm) | NN RMSE ± STDEV (mm) |
| --- | --- | --- | --- |
| Test 1: Vertical Compression | 4.37 ± 3.96 | 4.37 ± 3.96 | 6.2 ± 4.73 |
| Test 2: 6mm Diagonal +X | 3.61 ± 2.44 | 3.61 ± 2.44 | 7.71 ± 6.01 |
| Test 3: 6mm Diagonal -X | 6.05 ± 6.05 | 6.05 ± 6.05 | 6.73 ± 6.39 |
| Test 4 6mm Diagonal +Y | 23.62 ± 4.73 | 23.62 ± 4.73 | 11.8 ± 7.02 |
| Test 5: 6mm Diagonal -Y | 14.78 ± 6.05 | 14.78 ± 6.05 | 8.55 ± 5.6 |
| Test 6: 6mm Offset +X | 27.02 ± 5.89 | 27.02 ± 5.89 | 11.22 ± 6.67 |
| Test 7: 6mm Offset +Y | 21.08 ± 4.22 | 21.08 ± 4.22 | 11.57 ± 6.87 |

## V. HALL EFFECT DESIGN AND EVALUATION

### A. Proposed System

The input to the system consists of five Allegro A1326 hall effect sensors embedded in a plus configuration as shown in Fig. 1. These sensors are embedded directly into the bladder's base and are not mounted on a rigid substrate. The analog voltage output of these sensors then is fed through an amplification and filtering circuit as described in Fig. 5. Here, the signal has a first-order low-pass filter with a cutoff of 100Hz to reduce environmental noise, and an offset voltage is subtracted from the sensors. The A1326 sensor operates where 2.5 Volts is the bias point for zero measured field. For our purposes, only unipolar sensing is required, thus the offset voltage subtracts the bias point of the sensors to allow for the entire voltage range to be utilized for unipolar sensing. At the same time, a gain of 22 V/V is applied to the output to increase total sensitivity. The output from the amplifier stage is then sampled through a 10-bit, 2kHz analog to digital converter. These values are then used as inputs to a two-layer feed-forward neural network. Position in 3D (x, y, z) is then predicted and filtered using a moving average filter. MATLAB's regression fitter neural network was chosen for this architecture, as it provided adequate performance,

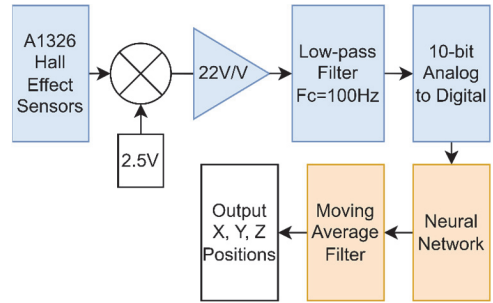

Fig. 5. Pipeline for processing Hall Effect sensor data and producing estimates of position.

portability to microcontrollers, and required no tweaking outside of the parameters explored here.

### B. Evaluation and Training Data Set Design

To evaluate the performance of the system, a series of bladder compressions were created that emulate those that would be experienced in a helmet impact. These include a vertical compression, diagonal compressions in each lateral axis, and offset compressions where the bladder is laterally manipulated to an offset before experiencing a vertical compression. A lateral offset of 6mm was chosen to represent a moderate deflection that would not permanently deform the bladder during repeated testing.

For training of the system, two methods were utilized. On initial approach of the problem, we chose to use a spiral trajectory, Fig. 3, that attempted to cover the cylindrical region that the bladder can deform within. Upon further investigation, it was discovered that although this pattern had a moderate distribution of points in the z-axis, it did not perform well in vertical compression tests. To overcome this, a new type of training trajectory that consists of a series of vertical compressions distributed through the cylindrical workspace was implemented, Fig. 4. After each vertical compression there is a diagonal path to the top of the next compression. There is a potential for bias where all the diagonal return paths face the same axis direction – for instance the positive X direction. To compensate for this, two trajectories were created where the first's return paths are in the X direction, and the second's return paths in the Y.

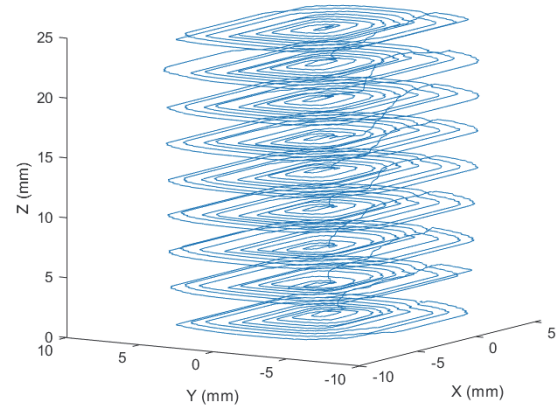

Fig. 3. Spiral path training set.

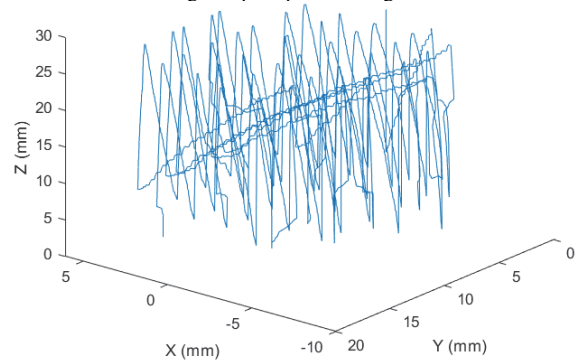

Fig. 4. Vertical compression path training set.

Table 2. Comparing RMSE for the spiral training set and the full training set (e.g., spiral + vertical).

| | Spiral Training RMSE ± STDEV (mm) [x, y, z] | Delta RMSE (mm) to Full Training Set. |
|---|---|---|
| Test 1: Vertical Compression | [0.76±0.63, 0.47±0.40, 0.44±0.35] | [0.56, 0.22, 0.31] |
| Test 2: 6mm Diagonal +X | [0.41±0.32, 0.42±0.33, 0.17±0.13] | [0.19, 0.17, 0.07] |
| Test 3: 6mm Diagonal -X | [0.30±0.49, 0.49±0.41, 0.07±0.06] | [0.08, 0.23, -.05] |
| Test 4 6mm Diagonal +Y | [0.69±0.67, 0.85±0.26, 0.13±0.11] | [0.14, 0.09, -0.02] |
| Test 5: 6mm Diagonal -Y | [0.82±0.75, 0.98±0.41, 0.14±0.11] | [0.15, 0.25, 0.00] |
| Test 6: 6mm Offset +X | [0.68±0.67, 0.75±0.44, 0.24±0.13] | [0.28, 0.68, 0.51] |
| Test 7: 6mm Offset +Y | [0.90±0.80, 0.80±0.36, 0.18±0.11] | [0.20, 0.27, 0.12] |

To compare the spiral training set with the more complete training set consisting of the spiral and vertical compression sets, two identical networks were trained on each set. These networks consisted of 15 hidden neurons and a 20-sample filter buffer. These values are derived as the best performance in the next sections. The performance of the network through each trial set is shown in Table 3 and the difference in error to the network trained with the full data set is shown in column 3. On average, the spiral training set produced 47% more error in the X axis, 58% more error in the Y axis, and 66% more error in the Z axis when compared to the full training set.

### C. Network Size Considerations

In this neural network architecture, the number of hidden neurons in the first layer is variable to compensate for over or under-fitting to training data. In addition to high neuron counts over-fitting data, it also adds computational complexity. To evaluate the effects of neuron count in this system, five different networks were trained on identical full training sets as defined in Sec V.B, with 20 sample output filter buffers. These networks ranged from 5 neurons to 30 neurons. Each network's performance was evaluated on all 8 test sets, and the RMSE of all trials was averaged to derive an aggregate RMSE of the network, Table 3. In these data sets, the error decreases from 5 neurons to 10 neurons to 15 neurons, but then ceases to reduce error significantly. For this reason, we selected 15 neurons as our best performance network.

### D. Output Filter Considerations

During development of the HE sensing system, it was noticed that the output of the NN was somewhat noisy, but this noise was centered around the correct prediction. For this reason, a moving average filter was implemented and added to the output of the NN. To derive the optimal buffer length of the filter, the optimal 15 node network was evaluated with five different sized filter buffers against the full test set. These buffer sizes are 0 (unfiltered), 10, 20, 30, and 40. Buffer sizes beyond 40 were not evaluated due to the phase lag created as a result. The results are shown in Fig. 7. As can be seen from the plot, the error in the system dropped until the 20-sample filter, where error stagnated or increased. For this reason, the 20 sample-filter was selected as the optimal for this system.

### E. Angle Desensitization

While the bladders can compress and move in up to six dimensions of movement, only five are sensible with a radially symmetric cylindrical magnet. For the application in the Smart Helmet, however, we decided to only actively sense the three translational components. This poses a potential problem however, as slight angular deflections can change the measured magnetic field significantly. To compensate for small angle deflections, a desensitization training set was created. In this set, the training trajectory was modified such that the translational components were identical, but two additional angular deflections were introduced. One of these sets was at five degrees, and another at ten degrees. All three training sets (0, 5, and 10 degrees of Y-axis deflection) were then concatenated and used to train the network.

To evaluate the performance of the system during small-angle deflections, the test trajectories were also modified for five and ten degree deflection. The performance of the desensitized system for each angle is shown in Table 3. The desensitization process did increase error in all categories, including on 0-degree tilt tests. This is most pronounced in the Z axis, where RMSE nearly doubled. Similarly, the direction in which the tilting was occurring (the Y-axis) had the most pronounced error as the tilt was increased. In future work, a full 5D sensing system should be evaluated that more robustly handles complex angular disturbances.

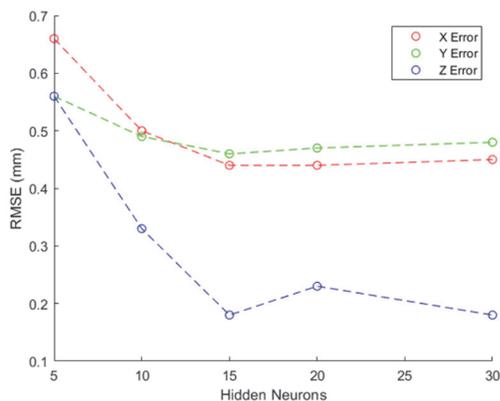

Fig. 6. Effect of network size on aggregate RMSE (mm) across Tests 1-8 for different NN hidden Neurons.

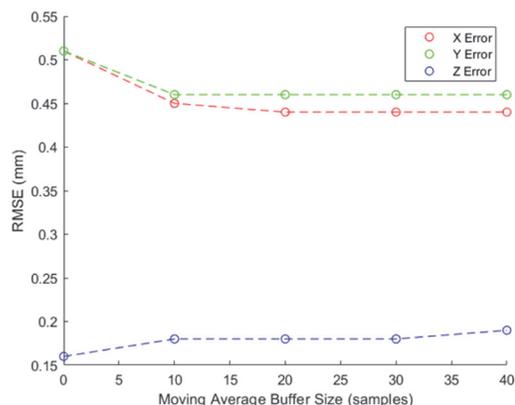

Fig. 7. Effect of moving average filter buffer on aggregate RMSE (mm) across Tests 1-8.

Table 3. RMSE with Neural Network trained on vertical, spiral, and angled training data by magnitude of bladder tilt in the Y-axis.

|  | 0 Degree Tilt | 5 Degree Tilt | 10 Degree Tilt |
|---|---|---|---|
|  | RMSE ± STDEV (mm) [x, y, z] | RMSE ± STDEV (mm) [x, y, z] | RMSE ± STDEV (mm) [x, y, z] |
| Test 1: Vertical Compression | [0.48±0.48, 0.87±0.55, 0.78±0.74] | [0.43±0.43, 1.34±0.32, 0.77±0.76] | [0.42±0.38, 1.27±0.32, 1.16±0.96] |
| Test 2: 6mm Diagonal +X | [0.48±0.43, 1.12±0.56, 1.15±1.13] | [0.50±0.45, 0.97±0.86, 0.84±0.77] | [0.68±0.61, 1.29±0.40, 0.96±0.73] |
| Test 3: 6mm Diagonal -X | [0.52±0.31, 0.30±0.23, 0.21±0.16] | [0.47±0.39, 1.42±0.36, 0.27±0.15] | [0.60±0.36, 1.30±0.31, 0.45±0.12] |
| Test 4: 6mm Diagonal +Y | [0.66±0.66, 0.43±0.32, 0.37±0.32] | [0.85±0.74, 0.96±0.51, 0.63±0.61] | [0.61±0.61, 1.22±0.70, 1.17±0.52] |
| Test 5: 6mm Diagonal -Y | [1.47±0.33, 0.65±0.35, 0.30±0.26] | [0.33±0.29, 1.58±0.34, 0.38±0.16] | [1.12±0.36, 1.50±0.32, 0.82±0.27] |
| Test 6: 6mm Offset +X | [1.16±0.27, 0.24±0.23, 0.13±0.11] | [0.39±0.27, 1.27±0.25, 0.15±0.15] | [1.50±0.35, 1.51±0.39, 0.65±0.16] |
| Test 7: 6mm Offset +Y | [0.93±0.26, 0.47±0.32, 0.27±0.24] | [0.60±0.25, 1.37±0.31, 0.70±0.31] | [1.84±0.28, 1.60±0.27, 0.90±0.39] |
| Average | [0.81±0.39, 0.58±0.37, 0.46±0.42] | [0.51±0.40, 1.27±0.42, 0.53±0.42] | [0.97±0.42, 1.38±0.39, 0.87±0.45] |

## VI. Conclusions and Future Work

While IR range finding is a simple and effective tool for localizing position in many applications, in our testing the performance of these solutions inside of a bladder can be heavily degraded. In vertical compression tests, the rangefinder is accurate to around 18mm at which point the noise in the readings increases substantially. This introduced noise makes fitting several types of models difficult, and the best RMSE achieved in this study was 4.7mm.

As an alternative, Hall effect localization of a magnet proved to be more accurate. This technique removed the need for line-of-sight to the top of the bladder and accounted for many of the non-linear effects inside of the bladder. In the same vertical compression test as the rangefinder, HE sensing achieved a RMSE of 0.44mm without accounting for angle perturbations, and 0.78 mm while accounting for angles up to 10 degrees. HE sensing also granted the ability to sense in the lateral directions in addition to compression.

For basic bladder classification schemes, IR range finding would most likely be adequate. The sensor is accurate enough to detect if the bladder is compressed and could offer some granularity as to maximum compression experienced. For more advanced control algorithms, especially those characterizing lateral forces, HE sensing would be more appropriate. Future revisions of the Smart Helmet could utilize both technologies. One possible scheme could utilize HE instrumented bladders in key points along the helmet and complement with rangefinders in between, potentially reducing overall complexity.

In future work, the repeatability of training across several bladders should be examined. Ideally, one trained network could be used for several bladders inside of one helmet. Several other system parameters need further exploration, such as the exact placement of Hall effect sensors, and optimization of the source magnetic field. Evaluation of the system in high-speed impacts should be considered as this may filtering. Interference between multiple bladders in proximity to each other should be characterized.

## Acknowledgements

The authors would like to thank Takara Truong, Tren Hirschi, and Alia Zaki for their preliminary results leading to this paper.